\documentclass{article} 
\usepackage{iclr2016_conference,times}
\usepackage{hyperref}
\usepackage{url}

\usepackage{amsmath}
\usepackage{amssymb}
\usepackage[linesnumbered,algoruled,boxed,lined]{algorithm2e}
\SetKwInOut{Parameter}{parameter}

\usepackage{fancybox}
\usepackage[margin=1in]{geometry}
\usepackage{graphicx}
\usepackage{color}

\title{QBDC: Query By Dropout Committee for training deep supervised architecture}

\author{Melanie Ducoffe \& Frederic Precioso\\
Univ. Nice Sophia Antipolis,\\
I3S, UMR UNS-CNRS 7271\\
06900 Sophia Antipolis, France\\
\texttt{\{ducoffe, precioso\}@i3s.unice.fr} \\
}

\begin{document}

\maketitle

\begin{abstract}
While the current trend is to increase the depth of neural networks to increase their performance, the size of their training database has to grow accordingly. We notice an emergence of tremendous databases, although providing labels to build a training set still remains a very expensive task. We tackle the problem of selecting the samples to be labelled in an online fashion. In this paper, we present an active learning strategy based on query by committee and dropout technique to train a Convolutional Neural Network (CNN). We derive a commmittee of \textit{partial CNNs}  resulting from batchwise dropout runs on the initial CNN. We evaluate our active learning strategy for CNN on MNIST benchmark, showing in particular that selecting less than 30 \% from the annotated database is enough to get similar error rate as using the full training set on MNIST. We also studied the robustness of our method against adversarial examples.
\end{abstract}

\section{Introduction}
Larger deep architectures fed with more data provide better results in error rate: This widely acknowledged idea has been confirmed all along the recent years when analyzing for instance the results at Imagenet Large Scale Visual Recognition Challenge \cite{ILSVRC15}. Indeed, in 2012, the winner was the SuperVision team \cite{dblp1728017} using a deep convolutional neural network with 60 million parameters and making a momentous breakthrough in the image classification task. The huge step forward from SuperVision team has deeply impacted the following contributions to ILSVRC after 2012. In 2014, Simonyan et al. \cite{CorrSimonyanZ14a} proposed also to use a CNN architecture from 11 up to 19 layers with 133 up to 144 million of parameters. In 2015, the winning team for the image classification context was GoogLeNet. They proposed a non-conventional CNN architecture allowing them to increase significantly both the depth and the width of the CNN while containing the computational cost.

The relation between the depth of the architecture, the required amount of training data and the final accuracy of the decision has not only been observed experimentally but it has also been explained in various papers. In their paper \cite{Bengio-nips-2006}, Bengio et al. explain clearly that complex decisions can be seen as highly-varying functions and that the decision making algorithm which intends to comprehend all these variations must be composed of many non-linearities. This specifics of deep architectures also impacts the compactness of the representation of highly-varying functions. In this same paper, they illustrate how deep architectures outperform shallow ones in terms of number of computational units required and thus training samples, in order to represent a given function. Their examples highlight even the differences in representation compactness between deep architectures, with respect to the problem.

Owing the considerable amount of parameters involved which needs to be learnt the training set needs to be huge as well. Furthermore, in order to cover the dispersion of the input distribution, strategies to extend the training set have emerged. For instance, when deep networks have been rolled out at the 2012 ImageNet challenge, they have been accompanied with image pre-processings to enlarge the training set \cite{dblp1728017}. In 2013, for instance, Howard \cite{CorrHoward13} started from the SuperVision team winning approach of 2012 and increased the accuracy by $20\%$ with such techniques. Even on smaller datasets, such as MNIST, the accuracy was improved when methods included a training set extension process also based on image transformations \cite{lecun1998gradient}.

Despite these techniques provide a better sampling of the target manifold, despite the \textit{``expressive power of deep architectures''} \cite{Bengio-2011} and despite their generalization power \cite{BengioTPAMI2013}, the work of Szegedy et al. \cite{szegedy2013intriguing} demonstrated this was not enough.

When considering the difficulty and the cost to gather relevant annotations for Challenges such as ImageNet \cite{ILSVRC15}, the need for methods requiring smaller training sets is increasing.

To this purpose, semi-supervised techniques have seen a recently increase into the recent publications. In deep learning especially,
research works are mixing unsupervised pretraining on the full training set with supervised classification on only a subset of it. While these techniques
has proven their efficiency with state of the art results, it remains that their performance depends, on some level, on the choice of the annotated subset.

Our work focuses on the selection of a better subset to be annotated for training, exploiting the theory of committee decisions.
We propose an adaptation of Query-By-Committee strategy (QBC) to deep learning. Indeed the huge number of parameters in a deep architecture prevents us from training a committee of deep networks.
Instead, we train a full Convolutional Neural Network (CNN) by selecting the training samples with a committee of partial CNNs based on batchwise dropout on the current full CNN, reducing the computational cost of the standard QBC technique.

\section{Related work}
Several learning strategies have been proposed to tackle the problem of selecting the best training samples in order to optimize the computational cost of training phase : Query-By-Committee, Semi-supervised methods, active learning strategies, and more recently active dropout deep architectures. 
\subsection{query by committee}
The first algorithm based on Query By Committee (QBC) strategy has been proposed by Seung et al. \cite{Seung1992}. They proved two important results: first, the generalization error (denoted $\epsilon_g$ in the original article) of a linear classifier for random training samples behaves as the inverse power law, $\epsilon_g(\alpha) \sim 1/\alpha$ where $\alpha = P/N$ with $P$ the number of training samples considered and $N$ the dimension of the input space; second, the generalization error of a linear classifier for training samples selectect through a query by committee strategy, scales like $\epsilon_g(\alpha) \sim e^{-\alpha I}$ with the constant decay is given by the information gain $I$. Later Freund et al. \cite{Freund1997} proved that this property hold for a more general class of learning problems.

In this paper, we consider a batch active learning based on Query-By-Committee.

\subsection{semi supervised learning}

Recently some works have been focusing on how to combine both unsupervised and supervised learning at the same time 
from a restricted annotated training set. Those semi supervised technique adapted for deep architectures are leaveraging
the size of the annotated database while being competitive with convolutional networks trained on the full annotated training set like done
in \cite{rasmus2015semi} or \cite{kingma2014semi}. However those methods are mostly based on generative networks \cite{kingma2013auto}, which are at the present time adapted to a restricted
area of inputs like images (\cite{kingma2014semi}) or texts( \cite{graves2013generating}).

\subsection{active learning for deep architectures}

Active learning has been first adapted recently to deep architectures for sentiment classification. In \cite{zhou2010active}, EZhou et al. proposed
an Active Deep Network (ADN) to select the most relevant reviews to constitute the annotated training set for their semi supervised tasks.
First they pretrain a Restricted Boltzmann Machine and use active learning on top of it : they ask the labels of a random unnatotated set of examples
and selects the ones which gets the least confidence on their label.

\subsection{Active Deep Dropout}
Simultaneous and independent work by \cite{gammelsaeterCommittee} also considers doing query by committee by applying dropout
on a standart Multi Layer Perceptron (MLP) to form a committee. To outperform the MLP accuracy, the first layers came from a pretrained DBN whose weights are
reused each time the MLP is reinitialized after increasing the labelled training set.
While their algorithm resembles ours, it differs in three main aspects wich make the difference on performances :
\begin{itemize}
 \item The author uses dropout to build the committee, making it not usable for convolutional architecture.
 \item The committee adds only one sample at a time, creating unbalanced weight among the training samples.
 \item The committee is built using dropout only with no rectification of its members. This may lead to a too rough approximation
 of the version space where the MLP lies. To solve this issue we provide some correction to improve the accuracy of the models in the committee.
\end{itemize}

\section{System description}

This section describes the design of our systems which aims at building a training set for a predetermined deep architecture. We first describe the selection criterion
our system builds on and then our method to build the committee on which our selection relies on.
\subsection{Selection criterion}

Out method consists in building a committee of deep architectures which selects among a subset of unlabelled samples, then pick the most
relevant ones to be labelled by an oracle and add them to the training set.
Between two selections of new samples, a deep network is train on the labelled training set using backpropagation until \textit{early stopping}.
For the sake of clarity, we denote by \textit{full network} the deep architecture trained on the labelled training set and \textit{partial network} a model of the committee.

Starting from a fixed architecture, its weights and biases are learnt so as to minimize the error of prediction on an independent validation set. We train initially the full network using a random annotated subset of the training samples.
It has experimentally been shown \cite{saxe2011random} that CNN initialized with random weights already hold some capability of discriminating classes (up to a certain accuracy of course) so that the initial annotated subset might be left empty, however we have chosen to start from an initial random subset of the training set to converge faster to relevant samples and benefit from the sample selection of our committee based approach from the first epochs.
After several epochs, the performance of this initial network does not increase anymore on the validation set. In order to improve the accuracy, we then inject a new minibatch of samples. Note that adding a minibatch of examples differs from \cite{gammelsaeterCommittee} in that add one sample at a time to the training set. Furthermore, adding minibatches of samples fosters gpu parallelization for training deep architectures, additionally to balancing every sample in the training set when applying mean gradient descent.

We hereby select a random subset of K samples from the rest of the training set.
A committee of deep networks assigns then a score to each sample from the selected minibatch, this score being related to the impact of the considered sample on the training of the deep architecture.
Denoting by \textit{batch\_size} the number of samples in a minibatch, the labels of the \textit{batch\_size} samples having the greatest score among the K samples  are then requesting. 
The score of a sample corresponds to the number of models in the committee whose prediction differs from the majority prediction among the committee members. The quality of the score is related to both the size of the committee and the correlation between the models \cite{melvilleicml04}.

 \begin{figure}
 \begin{center}
 \label{fig:QBDC}
 
  \caption{Query by dropout committee}
  
  \includegraphics[scale=0.1]{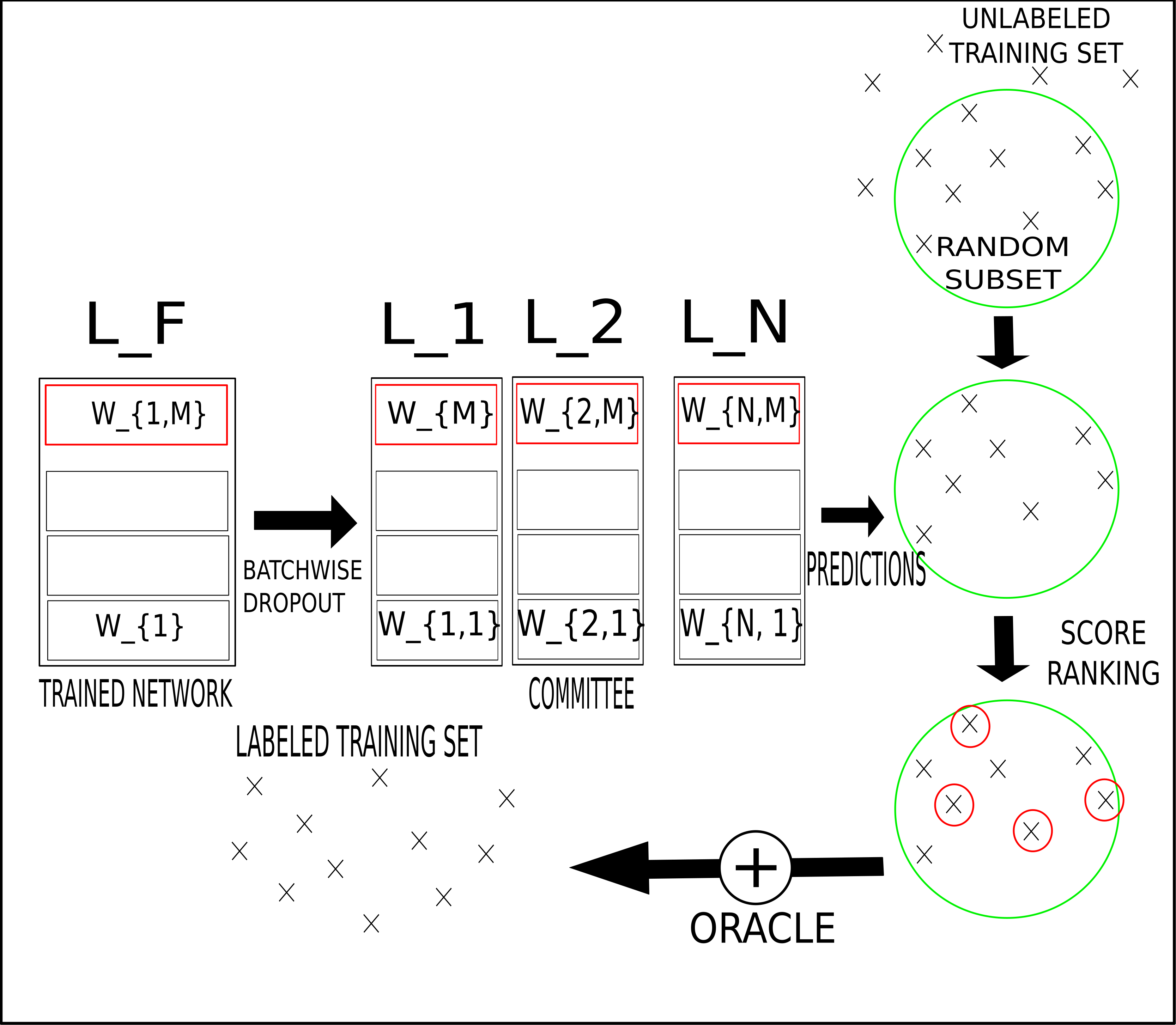}
  \end{center}
 \end{figure}

\subsection{Smart committee dropout}
The goal of the committee is to be representative of the version space where the current trained network lies. Indeed
it shows what a network, predicting correctly the annotated training set, could predict on samples from an uncovered part of the input distribution.
At first, as most part of the input distribution is not covered by the version space, most of the examples will be informative and the variance of their scores
will be relatively low. We can expect an increase of the variance as the knowledge of the manifold will improve with the size of the annotated training set, thus saving
a lot of the intervention of the human annotator. We illustrate our technique in figure~\ref{fig:QBDC}

Let us now detail how to form the committee. In the early papers describing active learning through committee selection, convergence and better result against randomness have been shown, first for perceptron-like learning functions \cite{Seung1992} and later for more general classes of learning problems \cite{Freund1997}.
However, for those results to hold, each model of our committee has to lie in the current version space defined by the annotated training set 
(which is the set of networks built from the same 'architecture' and making no mistake on the current training set).
It is worth noting already that this condition is not always feasible or suitable in the case of neural networks: 
\begin{enumerate}
 \item Indeed, we would need to apply backpropagation on each of the models of the committee to be effective on the training set. However, we are dealing with deep architectures, thus training several networks is not always computationally reasonnable.
 \item When training a network, we apply early stopping to avoid overfitting. Thus the current version space is not exactly defined on the training set but rather on a subset of it. When training a deep network, making no mistake on the training set is
 not always possible so we cannot assure to form a committee defined on the exact same version space than the full network's.
\end{enumerate}

To initiate a partial CNN while getting rid of the computation cost due to backpropagation, we apply batchwise dropout on our full network. 

The batchwise dropout \cite{graham2015efficient} is a version of dropout where we use a unique bernouilli mask to discard neurons for each sample in the minibatch.
Thus the batchwise dropout reduces the number of parameters in the architecture quadratically in the percentage of kept neurons.
Moreover dropout, when applied on convolutional layers removes neurons independently given the spatial locations. Whereas batchwise dropout
is spatially dependant, switching on or off filters so to discard neurons obtained through the same filter.
It is thus preserving the consistency in a CNN architecture. This technique is required to extend QBDC on CNN architecture which may be one of the reasons for the difference in the accuracy with \cite{gammelsaeterCommittee} that solely applied the dropout committee on MLP architectures.

In \cite{graham2015efficient}, a batchwise dropout committee has already been envisaged for future work in order to fasten the testing
while averaging the prediction on the full network \cite{baldi2014dropout}. The main advantage is to obtain a committee which the committee has the same architecture than the full network, with members having some zero constraints on several connections.
However because we are not using dropout while training the full network, some members of the committee
may have poor accuracy on the current training set, not being representative of the current version space.
In order to increase the accuracy of the committee, we retrain the last layer of every partial CNN via backpropagation.

Let consider the full network $N_F$ trained by backpropagation. We denote $P_{F} = softmax(a_{F})$ the output softmax with $a_F$ the vector of pre-softmax activations. Similarly, let $C_i$ be a partial network built with batchwise dropout on $N_F$ and $P_{i} = softmax(a_{i})$ the output softmax with $a_i$ the vector of pre-softmax activations. Suppose that $N_F$ is of depth $M$, every model $C_i$ is trained on its last layer $W_{i,M}$ so that its ouput $P_i$ is similar
to the true class of a sample $y_{true\_prediction}$. Every network in the process is trained to optimize the loss function cross entropy $\mathbb{H}$ as described in the following equations:
\begin{equation}
\label{equ:1}
 \mathbb{L}_F: \mathbb{L}( W \in N_F) = \mathbb{H}(y_{true\_prediction}, P_F)
\end{equation}
\begin{equation}
\label{equ:2}
 \mathbb{L}_i: \mathbb{L}( W_{i,M} ) = \mathbb{H}(y_{true\_prediction}, P_i)
\end{equation}

Equation~\ref{equ:1} refers to the standard cost function when training a network for supervised classification. On the other hand, the training for the committee in equation~\ref{equ:2} differs in that we modify only the last layers.
Thus we obtain an ensemble of smaller architectures, uncorrelated owing to the random selection of the nodes during dropout. The process is presented in algorithm~\ref{algo:QBDC}

\begin{algorithm}[H]
\caption{Query by committee \label{algo:QBDC}}
\SetKwInOut{Input}{Require}
\SetKwInOut{Output}{Prediction phase}
\SetKwInOut{Parameter}{Parameter}
\SetKwInOut{EndWhile}{End While}
\Input{$L^0$ set of initial labelled training examples (\textit{can be empty})}
\Input{$U^0$ set of initial labelled training examples}
\Input{n committee size}
\Input{$p_d$ dropout probability}
\Input{K number of samples analyzed by the committee}
\Parameter{hyperparameters value, batch size, architecture A}
j=0\\
\While{stop criterion}{
Train a network $N_F^j$ on $L^j$ built given the architecture A \\
Construct a committee $C^i$ of N classifiers $$C^i = DropoutSampling(N_F^j, p_d, n) $$ \\
\For{ i $\in [\mid 1, n \mid]$}{
Apply backpropagation given the loss function in equation~\ref{equ:2}\\
}
Select a random subset of K examples:  $S_{U^j} \subset U^j$  \\
\For{each $x_u\in S_{U^j}$}{
$\forall\;x_u$ Calculate the utility  $x_u$ based on the current committee utility $C^i$\\
Rank the examples in $S_{U^i}$ based on their utility\\
Select the batch\_size examples in $S_{U^j}$ with the highest rank: $S^j$\\
Ask an oracle to label examples in $S^j$\\
$U^{j+1} \gets U^j \setminus S^j$\\
$L^{j+1} \gets L^j\cup S^j$
}
}
return $N_F^j$
\end{algorithm}

\section{Experiments}

\subsection{Performance on benchmark datasets}
In this section, we analyze the performance of our active learning strategy on the benchmark MNIST.
\footnote{Code will be publicly available on the github account: https://github.com/mducoffe/ICLR\_2016}

\paragraph{MNIST:} We train a convolutional neural network with rectifier linear units for the full network $N_F$. The first two layers are convolutional layers with 20 and 40 filters of size (3,3) plus 2 by 2 pooling with no overlapping. We added on top of it two fully connected layers with 100 neurons in each layer.
We used RMSPProp \cite{tieleman2012lecture} with a learning rate of 0.001 and a decay rate of 0.9 with minibatches of size 200. The training set and the validation set are made of 50000 and 10000 samples respectively. Eventually when it comes to the hyperparameters of the active learning we picked up a committee of size 3 and init the annotated training set with 10 minibatches. To build the committee we apply a dropout rate of 0.5. The experiments have been repeated 5 times using different random seed and we provide the mean error and the minimal error (best accuracy over the 5 runs). 
We compare the performance with a random selection in the exact same state of hyperparameters. 

 \begin{figure}
 \begin{center}
 \label{fig:mnist_perf}
  \caption{Mean validation and testing error rate on MNIST for two active learning selections: QBDC and random selection. The performance
  are compared with the mean test error when training on the full network.}
  \includegraphics[scale=0.5]{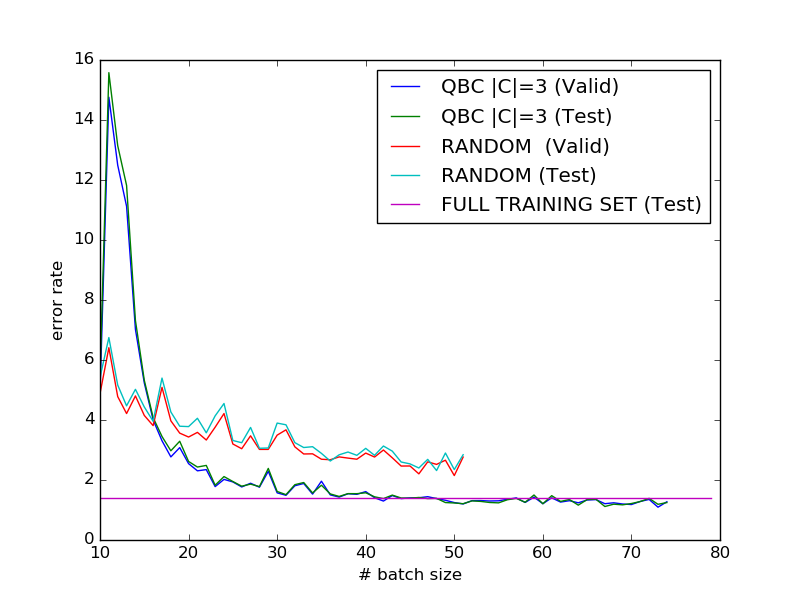}
   \end{center}
 \end{figure}
 
As we can see in figure~\ref{fig:mnist_perf} , the error rate when using active learning is always lower than when applying a random selection
, except for the first adds of samples. Indeed this peak results from a few networks having their accuracy worsening, while
the other experiments increase their accuracy. We make the assumption than this peak result from selection of outliers or adversarial examples.

Another observation is the time of convergence which is faster in the QBDC case : if we look at the slopes of both curve, they tend to diminish after 40 minibatches for both random and QBDC selection, with
an approximate 2\% of differences between their results. 
Although the average score when training on the standart full architecture (\textit{a network trained on the full labelled training set}) is of 1.38\%, which is higher than the state of the art on MNIST for CNN (0.95 \%, \cite{lecun1998gradient}),
the error rate on the QBDC full architecture (\textit{a network trained using only the samples selected with QBDC}) is competitive with the state of the art : we get an average error rate of 1.10 \%
while using less than 30 \% of the training set, as shown in table~\ref{tab:sample-table}

 \begin{table}[h]
\caption{Average and best error rates obtain using QBDC and random selection on MNIST using less than 30\% of the training set}
\label{tab:sample-table}
\begin{center}
\begin{tabular}{lll}
\multicolumn{1}{c}{\bf METHOD}  &\multicolumn{1}{c}{\bf MEAN error rate} &\multicolumn{1}{c}{\bf MIN error rate}
\\ \hline \\
QBDC & 1.10 & 0.99 \\
RANDOM & 2.13 & 1.78 \\
\end{tabular}
\end{center}
\end{table}


\subsection{Dropout short note}
To improve the initial performance when instantiating a committee, training the full network with dropout \cite{srivastava2014dropout} appears like an intuitive solution. Indeed when applying natural dropout proposed by Hinton et al. in \cite{srivastava2014dropout}, every possible model of the committee will be considered while training the full network. 
However, dropout is a technique designed for regularizing large networks with a certain amount of data. Consequently, when confronting to smaller network or restricted training set such as in MNIST, QBDC with dropout is not able to learn the features correctly, misleading the committee in its choices of new queries.

\begin{figure}
 \begin{center}
 \label{fig:mnist_dropout}
  \caption{Mean validation and testing error rate on MNIST for two active learning selection: QBDC and random selection. The performance
  are compared with the mean test error when training on the full network.}
  \includegraphics[scale=0.5]{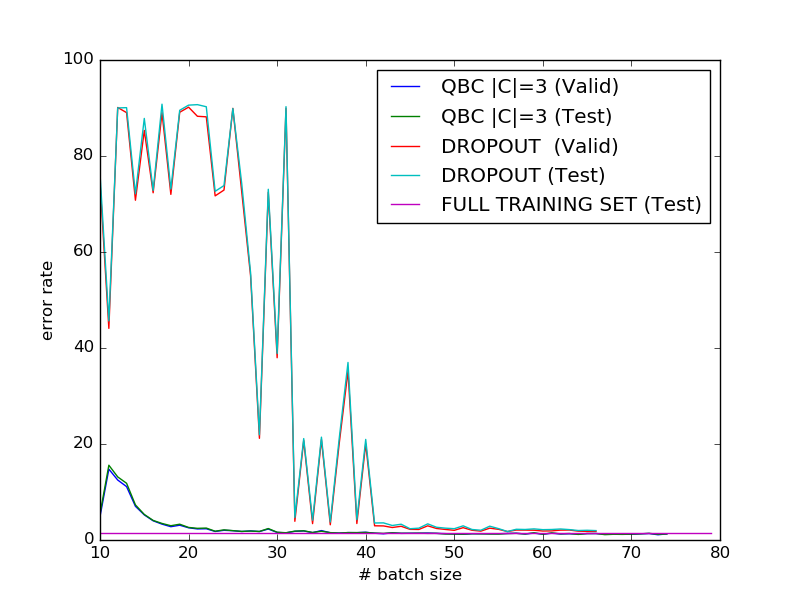}
   \end{center}
 \end{figure}

In the figure~\ref{fig:mnist_dropout}, QDBC with dropout using the exact same state of hyperparameters and a dropout rate of 0.5 is not generalizing
during the first epochs. We must wait after 30 minibatches (\textit{when the training database is sufficiently large}) than the full network
has learnt enough knowledge so that the committee starts selecting relevant samples. After 40 minibatches, the behavior of both error
rate for QDBC with and without dropout training tends to get similar, with a better accuracy of 0.4 \% for QBDC as shown in table~\ref{tab:sample-table_dropout}.
 
 \begin{table}[h]
\caption{Average and best error rates obtain using QBDC with and without dropout on MNIST using less than 30\% of the training set}
\label{tab:sample-table_dropout}
\begin{center}
\begin{tabular}{lll}
\multicolumn{1}{c}{\bf METHOD}  &\multicolumn{1}{c}{\bf MEAN error rate} &\multicolumn{1}{c}{\bf MIN error rate}
\\ \hline \\
QBDC & 1.10 & 0.99 \\
QBDC with dropout & 1.53 & 1.52 \\
\end{tabular}
\end{center}
\end{table}

\subsection{Adversary sensitive}
As any other machine learning, deep networks, especially CNNs are sensitive to adversarial examples, which are samples statistically diverging from the natural input distribution. Thus, every model of the committee may still be fooled by an adversarial example and select it to be labelled and added at the training examples.
Moreover, as pointed out in \cite{goodfellow2014explaining}, model averaging or dropout are not helping the network to be more robust from adversarial examples so that the committee decision strategy itself is still not preserving QBDC from considering and selecting adversarial examples.

Because we reduce the set of samples on which we train the full network, a natural question arising is: whether
considering only a subset of examples may alter the fitting to the input distribution and thus weakening the full network against
adversarial attacks.

To compare the robustness of the standart full classifier (\textit{ the classifier trained on the full training set}) and the QBDC full network  (\textit{a network trained using only the samples selected with
QBDC}, we compare
the number of adversarial examples found for different level of noise on 1000 samples on the test set of MNIST (\textit{over the 5 runs}). To generate efficiently approximate adversary we used  the \textit{fast gradient sign method} proposed by Goodfellow et al. in (\cite{goodfellow2014explaining}, see~\ref{adversary}).
We kept the exact same set of hyperparameters from the previous experiments.

\begin{table}[h]
\begin{center}
\caption{formula of the fast gradient sign method \cite{goodfellow2014explaining}}
\label{adversary}
\begin{itemize}
 \item $\Theta$ the parameters of the model
 \item $x$ the input of the model
 \item $y$ the targets associated with x
 \item $J(\Theta, x, y)$ the cost used to train the network
 \item $\eta$ the max norm constrained perturbation  \\
 \ovalbox{
 $\eta = \epsilon sign(\Delta_x J(\Theta, x, y))$
}
\end{itemize}
\end{center}
\end{table}

 \begin{figure}
 \begin{center}
 \label{fig:mnist_adv}
 
  \caption{Mean number of adversarial samples given the epsilon. We average on 1000 samples from the test set of MNIST}
  
  \includegraphics[scale=0.5]{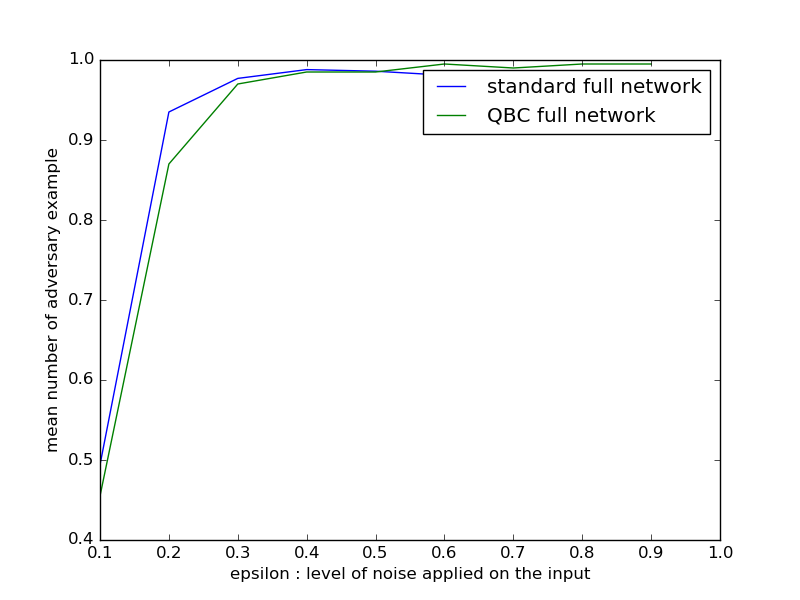}
   \end{center}
 \end{figure}

In figure~\ref{fig:mnist_adv}, we notice than our method tends to make the QBDC full network slightly more sensitive to adversarial
example. Indeed for low noise value ($epsilon = 0.1$), we found 4\% more adversarial examples on the QBDC full network. However these decrease is lower
compared to the amount of training samples discarded to train a QBDC full network (\textit{70\%}).

\section{Discussion}

Our QDBC method has demonstrated that using an automatic selection of the samples can reduce drastically
the size of the training set while preserving the accuracy for a fully supervised task. Preliminary results on ongoing
experiments are highlighting the usefulness of our method on more complicated distribution such as CIFAR10.
Eventually a drawback of our method, but shared by other active learning technique when applied to neural networks, is
the need to retrain the full network each time the committee adds a minibatch of new examples. We checked whether  the gain of accuracy when retraining the network
was low enough 

to avoid forgetting this step and save time. Doing so however has led to overfitting.
As a first tendency we have observed that down to some extreme case (committee of size 1 or 2) adding or
retriveving a few members in the commitee does not modify in average the accuracy.

As a future work, we would plan to analyze the samples selected by a committee and check for their relevancy.

\section{Conclusion}
In this paper, we have presented our own Query-By-Dropout-Committee (QBDC) for deep architectures. QBDC allows to train supervised deep networks on reduced annotated datasets by iteratively select the most relevant training samples. We have proved the efficiency of our approach: using only 30\% of the training set of MNIST we got close to state of the art performance while not degrading much the accuracy when confronting to adversarial examples.
These promising results will be extended to other machine learning techniques such as semi-supervised classification (ladder network) or regularization techniques such as batch normalization to handle huge datasets like ImageNet Challenge.

\section*{Acknowledgments}
We would like to thank the developpers of the frameworks Theano \cite{Bastien-Theano-2012}, Blocks and Fuel \cite{MerrienboerBDSW15} which we used for the experi-
ments.

\bibliography{iclr2016_conference}
\bibliographystyle{iclr2016_conference}

\end{document}